\documentclass{article}
\usepackage{spconf,amsmath,amssymb,graphicx,booktabs,diagbox,chngpage,float}
\restylefloat{table}

\title{GRAPH STRUCTURE BASED DATA AUGMENTATION METHOD}
\name{Kyung Geun Kim*, Byeong Tak Lee* \thanks{* These authors contributed equally to this work.}}
\address{VUNO Inc.}
\begin{document}

%
\maketitle
\begin{abstract}
In this paper, we propose a novel graph-based data augmentation method that can generally be applied to medical waveform data with graph structures. In the process of recording medical waveform data, such as electrocardiogram (ECG) or electroencephalogram (EEG), angular perturbations between the measurement leads exist due to discrepancies in lead positions. The data samples with large angular perturbations often cause inaccuracy in algorithmic prediction tasks. We design a graph-based data augmentation technique that exploits the inherent graph structures within the medical waveform data to improve both performance and robustness. In addition, we show that the performance gain from graph augmentation results from robustness by testing against adversarial attacks. Since the bases of performance gain are orthogonal, the graph augmentation can be used in conjunction with existing data augmentation techniques to further improve the final performance. We believe that our graph augmentation method opens up new possibilities to explore in data augmentation. 
\end{abstract}
\section{Introduction}
\label{sec:intro}
Data augmentation techniques have been popular method to reduce overfitting especially in the fields of computer vision~\cite{Shorten2019ASO, iwana2020empirical}. Applying augmentations to image data can be relatively intuitive since it is not very difficult to recognize the effect of an augmentation operation on the label change. This results from the fact that humans are naturally domain experts in image recognition tasks. However when dealing with medical waveform data, it is very difficult or even impossible to perceive the effect of an augmentation scheme on the label across the entire dataset. While only limited understandings of medical waveform data are available at human level, lead based measurement procedure introduces inherent graph structures in those data~\cite{JEKOVA201631}. From these observations, we propose a novel, nontrivial data augmentation scheme for medical waveform data that exploits these underlying graph structure.

The main ideas and experiments in this paper were developed using electrocardiogram (ECG) datasets as a concrete example, but it is not very difficult to see the generalizability of the suggested graph augmentation scheme to other medical waveform datasets such as electroencephalogram (EEG). In the process of recording 12-lead ECG signals, there exists an ideal location for each lead with fixed angular positions between them~\cite{WILSON194419}. These leads, as a whole, collectively measure the three dimensional electrical activities of the heart. Due to this fact, similarity between the measurements taken from the leads are higher if the leads are physically closer. Therefore, a graph structure naturally arises by the relative positions between the leads. We take advantage of these graph relation to propose an augmentation technique that introduce perturbations to a measurement taken from one lead using signals measured on the other leads. 

Existing augmentation methods in signal processing are normally no different than the methods used in computer vision. Specifically, the augmentations in the form of transformations in time or frequency domain as well as inserting random noises are normally applied~\cite{10.1007/978-3-319-73600-6_8,ko2017study}. While these augmentation methods are useful, they are incapable of providing perturbations that exploit the existing graph structure of the data. Since our method could generate augmentations that introduce perturbations with respect to the graph relations of lead measurements, we argue that our method could reduce the overfitting of a trained network to graph-structure-oriented noises. In addition, we show that the trained networks also become more robust to perturbations generated by adversarial attacks to demonstrate achieved robustness. Another advantage of our method is, because the augmentation is done in a dimension orthogonal to the traditional augmentation methods, they could have additive effect on the performance. We demonstrate the effectiveness of the proposed approach on diverse prediction tasks using three different datasets. 

The main contributions of our proposed methods include: (1) We introduce a new, generalizable graph augmentation technique that can be applied to any models across various tasks involving medical waveform datasets. (2) The proposed data augmentation method can enhance model robustness as well as overall performance. (3) We opened up a new possible direction in data augmentation technique to explore.
 
\section{METHODS}
\label{sec:methods}
    \begin{table*}[h!]
    \centering
        \resizebox{17.5cm}{!}{
            \begin{tabular}{l c c c c | c c c c | c c c c }
                \toprule
                & \multicolumn{4}{c}{Residual Network} & \multicolumn{4}{c}{Efficient Network} & \multicolumn{4}{c}{Densely Connected Network} \\
                \cline{2-13}
                & Base & RandAug & GraphAug & Both & Base & RandAug & GraphAug & Both & Base & RandAug & GraphAug & Both \\
                \hline
                CPSC  & 0.7896 & 0.8020 & 0.8032 & \textbf{0.8043} & 0.7570 & 0.7864 & 0.7750 & \textbf{0.7953} & 0.7869 & 0.7924  & 0.7932 & \textbf{0.8014} \\
                PTBXL & 0.7654  & 0.7675 & 0.7688 & \textbf{0.7718} & 0.7450  & 0.7543 & 0.7488 & \textbf{0.7566} & 0.7580 & 0.7623 & 0.7645 & \textbf{0.7659} \\
                SHAOXIN & 0.8718 & 0.8805 & 0.8854 & \textbf{0.8909} & 0.8605 & 0.8976 & 0.8814 & \textbf{0.9041} & 0.8556 & 0.8698 & 0.8727 & \textbf{0.8769}\\
                \toprule

            \end{tabular}
        }
        \caption{Performance of neural network models for each dataset trained according to different augmentation schemes}
        \label{table:1}
    \end{table*}

    \subsection{Graph Augmentation Design}
    \label{ssec:graphcon}
    In order to construct augmentation graph, we first examine how 12-lead ECG signals are measured and processed. The 12-lead ECG is defined by aVR, aVL, aVF, I, II, III and V1-V6 leads, which are measured by 6 precordial electrodes and 4 limb electrodes. The leads can be interpreted as real or virtual positions in a space to measure three dimensional electrical activities generated by the heart~\cite{bear2015forward}. Naturally, since the electrical potentials measured at two close locations must be similar, the signal similarity very much depends on the physical, relative locations of each lead. We used these similarity information to represent the graph structure between the 12 ECG leads. Specifically, average correlation (cross correlation at lag $k=0$) of the lead measurements are computed to construct the weighted adjacency matrix $\mathbf{A}$ as computed below:
    \begin{equation}
        \mathbf{A}_{ij} = \mathbb{E} \left[{\frac{\sum_{t=1}^{T} {(X^{(i)}_t - \mu^{(i)}) (X^{(j)}_t - \mu^{(j)})}}{\sigma^{(i)} \sigma^{(j)}}}\right], \ \ i \neq j
    \end{equation}
    where $\mathbf{X}^{(i)}, \mathbf{X}^{(j)}$ are the signal vectors of length $T$ measured at leads $i$ and $j$. For the case $i = j,\ \mathbf{A}_{ij} = 0$ is chosen.  
    
\begin{figure}[h!]
      \centering
      \includegraphics[width=8.5cm]{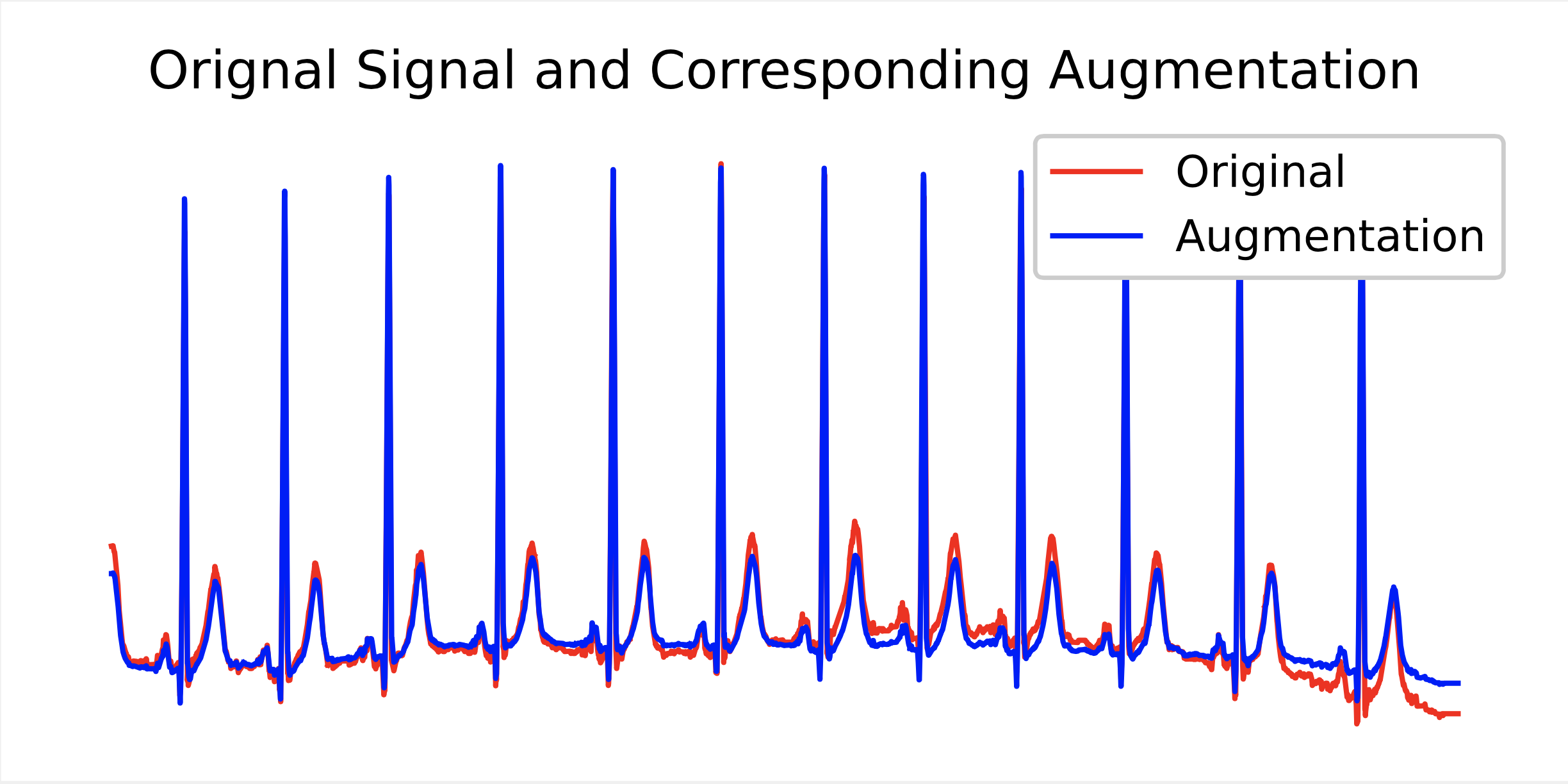}
        \caption{Comparison between an original signal and its corresponding augmentation}
        \label{figure:1}
    \end{figure}

    Using the graph defined by $\mathbf{A}$ as the augmentation graph, a linear combination of the other leads could be constructed for one specific lead $i$:
    \begin{equation}
        \mathbf{\widetilde{X}}^{(i)} = \sum_{j \neq i} {\mathbf{A}_{ij} \mathbf{X}^{(j)}}
    \end{equation}
    The constructed signal $\mathbf{\widetilde{X}}^{(i)}$ combines the signals measured from other leads according to how similar they are to lead $i$. Therefore, it is reasonable to think of $\mathbf{\widetilde{X}}^{(i)}$ as a graph structure induced augmentation. To illustrate the similarity, example of such augmented signal $\mathbf{\widetilde{X}}^{(i)}$ and original signal $\mathbf{X}^{(i)}$ is shown in Figure~\ref{figure:1}. 
    Using the augmented signal $\mathbf{\widetilde{X}}^{(i)}$, a convex combination between the original signal and the final augmentation at some lead $i$ is computed:
    \begin{equation}
        \mathbf{\widehat{X}}^{(i)} = (1-\lambda) \mathbf{X}^{(i)} + \lambda \mathbf{\widetilde{X}}^{(i)}, \ \ \lambda \in [0, 1]
    \end{equation}
    The augmentation is randomly applied with probability of $p$ where parameter $\lambda$ is drawn from a uniform distribution $\mathcal{U}(0, \alpha)$ every time $\mathbf{\widehat{X}}^{(i)}$ is computed. The augmentation parameters $p$ and $\alpha$ are hyperparameters to be tuned. 
    
    \subsection{Augmentation Combination Details}
    \label{ssec:strucdetil}
    In order to compare our augmentation method to other methods, we selected several popular existing augmentation techniques. The parameter combinations determining how each augmentation is applied are tuned using RandAugment~\cite{cubuk2020randaugment} for peak performance. In order to apply RandAugment~\cite{cubuk2020randaugment}, an intensity parameter shared across all of the augmentation schemes must be defined. Let $\gamma$ be an intensity parameter shared across every augmentation schemes. With $\gamma$, we applied described augmentations according to following rules. (1) Adding a Gaussian noise for every time point drawn i.i.d from $\mathcal{N}(0, \gamma)$~\cite{um2017data}. (2) Transformation in time domain where $\gamma$\% of the signal is randomly cut or zero-padded to be expanded or compressed using bilinear interpolation to fit to the original length~\cite{le2016data}. (3) Smoothing using weighted moving average window defined by Gaussian kernel with length, $l \in \{1, 2, 3, 4, 5\}$ which is chosen with equal probability according to $\gamma$~\cite{nguyen2020improving}. (4) Masking operation which $\gamma$\% of the signal is masked with zero starting at a position chosen uniformly random~\cite{park2019specaugment}. 
    
\begin{figure}[h!]
      \centering
      \centerline{\includegraphics[width=8.55cm]{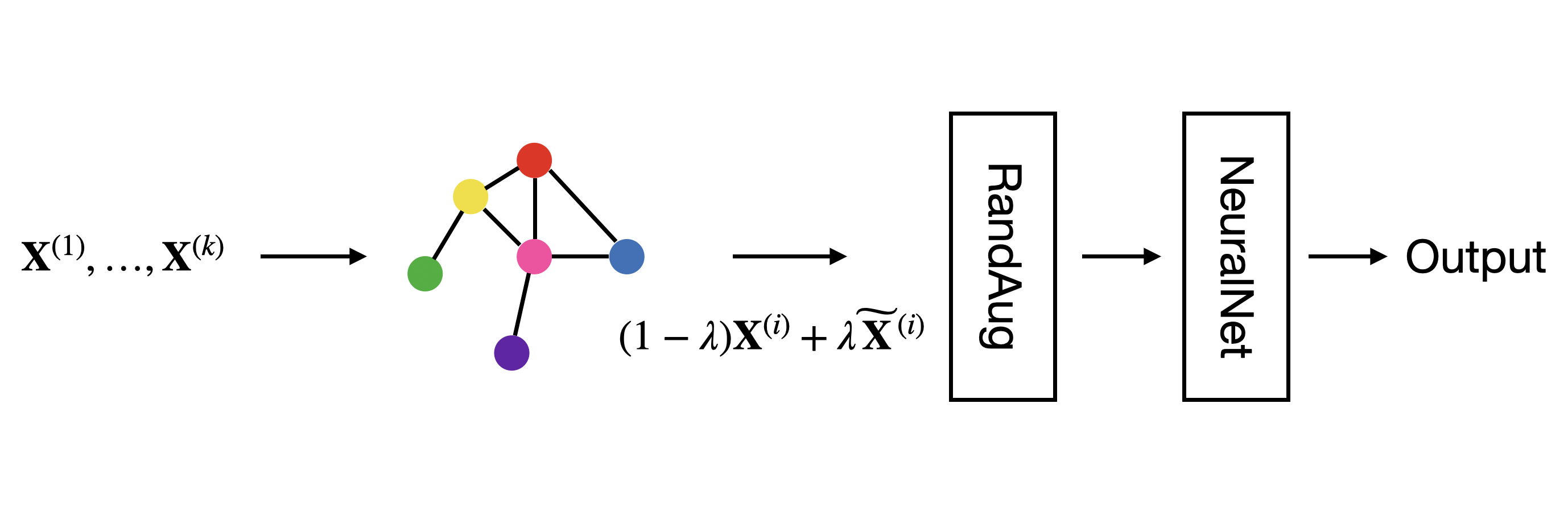}}
        \caption{Overall structure of augmentation module}
        \label{figure:2}
    \end{figure}

    When applying graph augmentation and traditional augmentations at same time, we first apply the graph augmentation followed by the traditional augmentations. The order of which augmentations are applied first is critical because of the following reasons. Traditional, lead-wise perturbations are done using different strategies meaning that if the graph augmentation is applied afterwards, a single trace of a waveform will contain mix of multiple augmentations at same time, possibly becoming too noisy. More importantly, the correlation values of the augmentation graph will no longer be valid after normal augmentations are applied. Therefore, applying traditional augmentations before the graph augmentation will have an adverse affect on the performance of the neural networks to be trained. The overall structure of augmentation module is presented in Figure~\ref{figure:2}. 
    
\section{EXPERIMENTS}
\label{sec:experiements}

    \subsection{Datasets and Models Description}
    \label{ssec:dataset}
    The experiments are carried out using three different open databases of The China Physiological Signal Challenge 2018 (CPSC)~\cite{liu2018open}, SHAOXING~\cite{zheng202012} and Physikalisch Technische Bundesanstalt-XL (PTBXL)~\cite{wagner2020ptb} which includes 7,166, 10,615 and 21,837 data points each. Each dataset includes labels for different types of arrhythmia. CPSC and PTBXL datasets contain labels fit for mult-label tasks for following arrhythmia detection: Atrial fibrillation, First-degree atrioventricular block, Left bundle brunch block, Right bundle brunch block, Premature atrial contraction, Premature ventricular contraction, ST-segment depression, and ST-segment elevated for CPSC and Normal ECG, Myocardial Infarction, ST/T Change, Conduction Disturbance and additionally, Hypertrophy for PTBXL. SHAOXING dataset contains labels fit for multi-class classification tasks of seven classes composed of Sinus Bradycardia, Sinus Rhythm, Atrial Fibrillation, Sinus Tachycardia, Atrial Flutter, Sinus Irregularity, and Supraventricular Tachycardia. The training, validation and testing sets are divided within the same datasets into 70\%, 15\%, and 15\%, respectively. 
    
    To investigate the efficacy of the proposed method, we tested on the modification of well known models: Residual Network (ResNet)~\cite{he2016deep}, Efficient Network (EfficientNet) ~\cite{tan2019efficientnet} and Densely Connected Network (DenseNet)~\cite{huang2017densely}. Since these networks are originally designed for low resolution image classification, we modified them for usage in 1D signals. For ResNet, we implemented a modification of ResNet that is popularly used for handling ECG data~\cite{hannun2019cardiologist}. To adjust for the length of the signal, the pooling layer of the last block is modified from 2 to 1. For EfficientNet, we implemented modified version of EfficientNet-B0. Specifically, 2-dimensional $N \times N$ filters are changed to 1-dimensional $N^2$ filters and stride operation is added at the 5th stage. For DenseNet, we set the growth rate to 12 and filter size to 16 for the entire network where each block consists of 6-bottleneck layers. The networks are trained using Adam optimizer with a learning rate of 0.001. We used dropout regularization with $p=0.1$ and batch size of 32. $l_2$ regularization with a coefficient of $10^{-5}$ is applied.

    

    \subsection{Performance Evaluation and Comparison}
    \label{ssec:eval}
    In order to evaluate the performance of our augmentation method, we first compared performance gain of graph augmentation against no augmentation as well as normal augmentations. We also show that the graph augmentation method can be applied on top of normal augmentations by showing that they have additive effect on the performance. The results of these experiments are summarized in Table~\ref{table:1} where each entries are F1 scores averaged over multiple repeated experiments. Although performance gain achieved from normal augmentations comes from optimal parameter combination tuned by RandAugment, graph augmentation and normal augmentations show similar performance gain on average across every model and dataset. Additionally, when graph augmentation and normal augmentations are applied at the same time, additional performance gain has been observed across every dataset and model. As results of these experiments, we arrived at the following conclusions. First, we show that using graph augmentation alone has a similar impact on network performance than using not only one, but multiple, well tuned normal augmentation methods. Second, we conclude that performance gain from graph augmentation is orthogonal to normal augmentations resulting in additive effect on performance. 
    
    \subsection{Augmentation Parameters Optimization}
    \label{ssec:augparamselect}
    Depending on how the augmentation parameters are tuned, they can have critical effects on the performance of the trained networks. For maximal gain in  performance, we have tuned these parameters for both normal and graph augmentations using RandAugment. We observed that, for the normal augmentations, following RandAugment scheme clearly outperforms hand searched parameter combinations.
    
\begin{figure}[h!]
      \centering
      \centerline{\includegraphics[width=9cm]{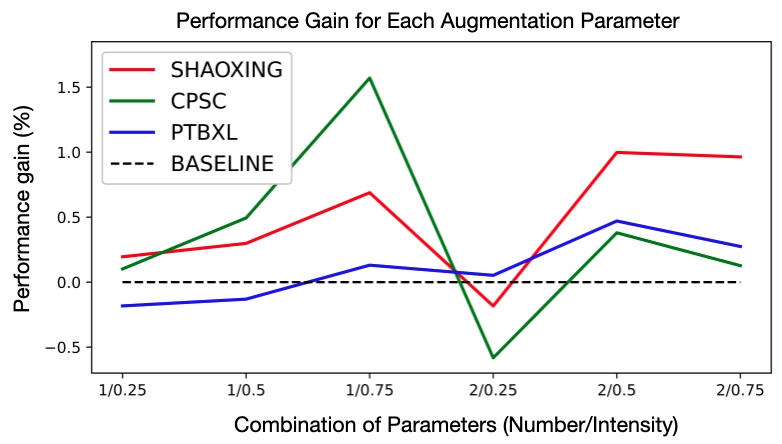}}
        \caption{Performance of ResNet trained with different augmentation parameter combinations selected by RandAugment}
        \label{figure:3}
    \end{figure}

    We have tried six different augmentation parameter combinations measuring the performance gain achieved for each combination. As the summery of performance gain for ResNet shows in Figure~\ref{figure:3}, performance gain achieved by augmentation methods are usually quite marginal. In addition, applying multiple augmentations on top of each other does not guarantee performance gains~\cite{cubuk2020randaugment}. However, since the performance gain resulting form graph augmentation is on par with RandAugment, we conclude that it can provide considerable amount of performance gain compared to other, single augmentation methods.
    
    \subsection{Robustness Effect of Graph Augmentation}
    \label{ssec:robustness}
    
    We hypothesized that the main reason of the performance gain achieved by graph augmentation method lies within the network's robustness with respect to graph structure induced perturbations. To elaborate, if the network trained with graph augmentation is robust to these perturbations, it will make more correct predictions on the perturbed data samples when tested. To show the network robustness against graph perturbations, testing datasets with large graph structure based perturbations are required. An ideal way to build such test set is to have lead-wise, positional perturbations at the measurement level. Since this is intractable, we instead constructed perturbed data samples using adversarial attack method~\cite{madry2017towards} which finds an (sub) optimal perturbation to change the label of the data as best as possible. Therefore by demonstrating adversarial robustness, we believe sufficient evidences are provided to explain the performance gain achieved by graph augmentation. In addition, we believe that the graph augmentation and normal augmentation schemes have additive effect on performance for the following reason. In contrast to graph augmentation, the input data generated by normal augmentations introduces perturbations that are limited to each \textit{waveform} rather than the relations between. As a result, the normal augmentation reduces overfitting related to noises within each waveform signals which is independent from the effect of the graph augmentation. For this reason, the two different augmentation methods implement robustness that are orthogonal to each other having additive effect when applied at same time. 
    
    \begin{figure}[h!]
      \centering
      \centerline{\includegraphics[width=9cm]{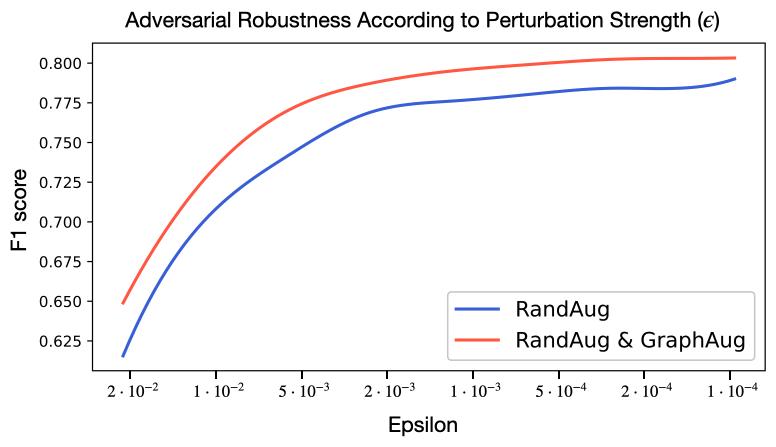}}
        \caption{Adversarial robustness of ResNets trained using RandAugment and using both RangAugment and graph augmentation according to different perturbation strength ($\epsilon$)}
        \label{figure:4}
    \end{figure}
    
    In the robustness experiments, we used ResNet trained on CPSC dataset evaluating the F1 scores according to the perturbation strength $\epsilon$. We evaluated the effect of the graph augmentation by comparing the F1 score decay of the network trained with only RandAugment and the network trained with graph augmentation added on top. As shown in Figure \ref{figure:4}, the network trained with both augmentations is more robust by a significant margin for every $\epsilon$ tested. Thus, we conclude that the graph augmentation can offer robustness with respect to the adversely constructed perturbations.

\section{CONCLUSION}
\label{sec:conclusion}
In this work, we propose a novel data augmentation method called graph augmentation that exploits underlying graph structures of medical waveform data. By using average correlations between the signals recorded from each lead, approximations of recordings from 12-lead ECG signals are constructed. We used these with the original signal, and taking a convex combination, we successfully introduced augmentations that has minimal effects on the labels of each data samples. We show the effect of graph augmentation method in three different open datasets and in three different neural network models. In addition, we also show the effect of the graph augmentation on the network's robustness with respect to adversarial attacks. A critical extension we would like to show in our future work is how our method generalizes to other data formats such as electroencephalograms. As the direction of our future work suggests, we have introduced a new direction to explore in data augmentation methods for datasets that have underlying graph structures. 


\bibliographystyle{IEEEbib}
\bibliography{refs}

\begin{thebibliography}{10}

\bibitem{Shorten2019ASO}
Connor Shorten and T.~Khoshgoftaar,
\newblock ``A survey on image data augmentation for deep learning,''
\newblock {\em Journal of Big Data}, vol. 6, pp. 1--48, 2019.

\bibitem{iwana2020empirical}
Brian~Kenji Iwana and Seiichi Uchida,
\newblock ``An empirical survey of data augmentation for time series
  classification with neural networks,''
\newblock {\em arXiv preprint arXiv:2007.15951}, 2020.

\bibitem{JEKOVA201631}
Irena Jekova, Vessela Krasteva, Remo Leber, Ramun Schmid, Raphael Twerenbold,
  Christian Müller, Tobias Reichlin, and Roger Abächerli,
\newblock ``Inter-lead correlation analysis for automated detection of cable
  reversals in 12/16-lead ecg,''
\newblock {\em Computer Methods and Programs in Biomedicine}, vol. 134, pp. 31
  -- 41, 2016.

\bibitem{WILSON194419}
Frank~N. Wilson, Franklin~D. Johnston, Francis~F. Rosenbaum, Herman Erlanger,
  Charles~E. Kossmann, Hans Hecht, Nelson Cotrim, Roberto~Menezes {de
  Oliveira}, Roberto Scarsi, and Paul~S. Barker,
\newblock ``The precordial electrocardiogram,''
\newblock {\em American Heart Journal}, vol. 27, no. 1, pp. 19 -- 85, 1944.

\bibitem{10.1007/978-3-319-73600-6_8}
Fang Wang, Sheng-hua Zhong, Jianfeng Peng, Jianmin Jiang, and Yan Liu,
\newblock ``Data augmentation for eeg-based emotion recognition with deep
  convolutional neural networks,''
\newblock in {\em MultiMedia Modeling}, Klaus Schoeffmann, Thanarat~H.
  Chalidabhongse, Chong~Wah Ngo, Supavadee Aramvith, Noel~E. O'Connor, Yo-Sung
  Ho, Moncef Gabbouj, and Ahmed Elgammal, Eds., Cham, 2018, pp. 82--93,
  Springer International Publishing.

\bibitem{ko2017study}
Tom Ko, Vijayaditya Peddinti, Daniel Povey, Michael~L Seltzer, and Sanjeev
  Khudanpur,
\newblock ``A study on data augmentation of reverberant speech for robust
  speech recognition,''
\newblock in {\em 2017 IEEE International Conference on Acoustics, Speech and
  Signal Processing (ICASSP)}. IEEE, 2017, pp. 5220--5224.

\bibitem{bear2015forward}
Laura~R Bear, Leo~K Cheng, Ian~J LeGrice, Gregory~B Sands, Nigel~A Lever,
  David~J Paterson, and Bruce~H Smaill,
\newblock ``Forward problem of electrocardiography: is it solved?,''
\newblock {\em Circulation: Arrhythmia and Electrophysiology}, vol. 8, no. 3,
  pp. 677--684, 2015.

\bibitem{cubuk2020randaugment}
Ekin~D Cubuk, Barret Zoph, Jonathon Shlens, and Quoc~V Le,
\newblock ``Randaugment: Practical automated data augmentation with a reduced
  search space,''
\newblock in {\em Proceedings of the IEEE/CVF Conference on Computer Vision and
  Pattern Recognition Workshops}, 2020, pp. 702--703.

\bibitem{um2017data}
Terry~T Um, Franz~MJ Pfister, Daniel Pichler, Satoshi Endo, Muriel Lang, Sandra
  Hirche, Urban Fietzek, and Dana Kuli{\'c},
\newblock ``Data augmentation of wearable sensor data for parkinson’s disease
  monitoring using convolutional neural networks,''
\newblock in {\em Proceedings of the 19th ACM International Conference on
  Multimodal Interaction}, 2017, pp. 216--220.

\bibitem{le2016data}
Arthur Le~Guennec, Simon Malinowski, and Romain Tavenard,
\newblock ``Data augmentation for time series classification using
  convolutional neural networks,''
\newblock 2016.

\bibitem{nguyen2020improving}
Thai-Son Nguyen, Sebastian St{\"u}ker, Jan Niehues, and Alex Waibel,
\newblock ``Improving sequence-to-sequence speech recognition training with
  on-the-fly data augmentation,''
\newblock in {\em ICASSP 2020-2020 IEEE International Conference on Acoustics,
  Speech and Signal Processing (ICASSP)}. IEEE, 2020, pp. 7689--7693.

\bibitem{park2019specaugment}
Daniel~S Park, William Chan, Yu~Zhang, Chung-Cheng Chiu, Barret Zoph, Ekin~D
  Cubuk, and Quoc~V Le,
\newblock ``Specaugment: A simple data augmentation method for automatic speech
  recognition,''
\newblock {\em arXiv preprint arXiv:1904.08779}, 2019.

\bibitem{liu2018open}
Feifei Liu et~al.,
\newblock ``An open access database for evaluating the algorithms of
  electrocardiogram rhythm and morphology abnormality detection,''
\newblock {\em Journal of Medical Imaging and Health Informatics}, vol. 8, no.
  7, pp. 1368--1373, 2018.

\bibitem{zheng202012}
Jianwei Zheng, Jianming Zhang, Sidy Danioko, Hai Yao, Hangyuan Guo, and Cyril
  Rakovski,
\newblock ``A 12-lead electrocardiogram database for arrhythmia research
  covering more than 10,000 patients,''
\newblock {\em Scientific Data}, vol. 7, no. 1, pp. 1--8, 2020.

\bibitem{wagner2020ptb}
Patrick Wagner, Nils Strodthoff, Ralf-Dieter Bousseljot, Dieter Kreiseler,
  Fatima~I Lunze, Wojciech Samek, and Tobias Schaeffter,
\newblock ``Ptb-xl, a large publicly available electrocardiography dataset,''
\newblock {\em Scientific Data}, vol. 7, no. 1, pp. 1--15, 2020.

\bibitem{he2016deep}
Kaiming He, Xiangyu Zhang, Shaoqing Ren, and Jian Sun,
\newblock ``Deep residual learning for image recognition,''
\newblock in {\em Proceedings of the IEEE conference on computer vision and
  pattern recognition}, 2016, pp. 770--778.

\bibitem{tan2019efficientnet}
Mingxing Tan and Quoc~V Le,
\newblock ``Efficientnet: Rethinking model scaling for convolutional neural
  networks,''
\newblock {\em arXiv preprint arXiv:1905.11946}, 2019.

\bibitem{huang2017densely}
Gao Huang, Zhuang Liu, Laurens Van Der~Maaten, and Kilian~Q Weinberger,
\newblock ``Densely connected convolutional networks,''
\newblock in {\em Proceedings of the IEEE conference on computer vision and
  pattern recognition}, 2017, pp. 4700--4708.

\bibitem{hannun2019cardiologist}
Awni~Y Hannun, Pranav Rajpurkar, Masoumeh Haghpanahi, Geoffrey~H Tison, Codie
  Bourn, Mintu~P Turakhia, and Andrew~Y Ng,
\newblock ``Cardiologist-level arrhythmia detection and classification in
  ambulatory electrocardiograms using a deep neural network,''
\newblock {\em Nature medicine}, vol. 25, no. 1, pp. 65, 2019.

\bibitem{madry2017towards}
Aleksander Madry, Aleksandar Makelov, Ludwig Schmidt, Dimitris Tsipras, and
  Adrian Vladu,
\newblock ``Towards deep learning models resistant to adversarial attacks,''
\newblock {\em arXiv preprint arXiv:1706.06083}, 2017.

\end{thebibliography}

\end{document}